\documentclass[journal]{IEEEtran}

\usepackage{hyperref}
\usepackage{cite}
\usepackage[pdftex]{graphicx}
\usepackage{amsmath}
\usepackage{amsfonts}
\usepackage[caption=false,font=footnotesize]{subfig}
\usepackage{url}
\usepackage{multirow}

\DeclareMathOperator*{\argmin}{argmin}

\begin{document}
\title{\huge \bf ORB-SLAM2: an Open-Source SLAM System for \\Monocular, Stereo and RGB-D Cameras}

\author{Ra\'ul Mur-Artal and Juan D. Tard\'os%
\thanks{This work was supported by the Spanish government under Project DPI2015-67275, the Arag\'on regional governmnet under Project DGA T04-FSE and the Ministerio de Educaci\'on Scholarship FPU13/04175.}%
\thanks{R. Mur-Artal was with the Instituto de Investigaci\'on en Ingenier\'ia de Arag\'on (I3A), Universidad de Zaragoza, 50018 Zaragoza, Spain, until January 2017. He is currently with Oculus Research, Redmond, WA 98052 USA (e-mail:
raul.murartal@oculus.com).}%
\thanks{J. D. Tard\'os is with the Instituto de Investigaci\'on en Ingenier\'ia de Arag\'on (I3A), Universidad de Zaragoza, 50018 Zaragoza, Spain (e-mail:
tardos@unizar.es).}}

\thispagestyle{empty} 
\onecolumn 
 
\begin{center} 
\noindent 
 
This paper has been accepted for publication in \emph{IEEE Transactions on Robotics}. 
 
\vspace{2em} 
 
DOI: \href{https://doi.org/10.1109/TRO.2017.2705103}{10.1109/TRO.2017.2705103} 
 
IEEE Xplore: \url{http://ieeexplore.ieee.org/document/7946260/} 
\end{center} 
\vspace{3em} 
 
\copyright2017 IEEE. 
Personal use of this material is permitted. Permission  
from IEEE must be obtained for all other uses, in any current or  
future media, including reprinting 
/republishing this material for  
advertising or promotional purposes, 
creating new collective works, for resale or  
redistribution to servers or lists, or reuse of any copyrighted  
component of this work in other works. 
 
\twocolumn

\setcounter{page}{1}

\maketitle

\begin{abstract}
We present ORB-SLAM2 a complete SLAM
system for monocular,  stereo and RGB-D cameras, including map reuse, loop
closing and relocalization capabilities. The system works in real-time on standard
CPUs in a wide variety of environments
from small hand-held indoors sequences, to drones flying in
industrial environments and cars driving around a city. Our back-end
based on bundle adjustment with monocular and stereo observations
allows for accurate trajectory estimation with metric scale. 
Our system includes a lightweight localization mode that leverages
visual odometry tracks for unmapped regions and matches to map points
that allow for zero-drift localization. The evaluation on 29
popular public sequences shows that our method achieves state-of-the-art accuracy, being in most cases the most accurate SLAM solution.
We publish the source code, not only for the benefit of the SLAM
community, but with the aim of being an out-of-the-box SLAM solution
for researchers in other fields.
\end{abstract}

\section{Introduction}
Simultaneous Localization and Mapping (SLAM) has been a hot research topic in the last two decades in the Computer Vision and Robotics communities, and has recently
attracted the attention of high-technological companies. SLAM techniques build a map of an unknown environment and localize the sensor in the map with a strong focus on real-time 
operation. Among the different sensor modalities, cameras are cheap and provide rich information of the environment that allows for robust and accurate place recognition. 
Therefore Visual SLAM solutions, where the main sensor is a camera, are of major interest nowadays.
Place recognition is a key module of a SLAM system to close loops (i.e. detect when the sensor returns to a mapped area and correct the accumulated error in exploration) 
and to relocalize the camera after a tracking failure, due to occlusion or aggressive motion, or at system re-initialization. 

Visual SLAM can be performed by using just a monocular camera, which is the cheapest and smallest sensor setup. 
However as depth is not observable from just one camera, the scale of the map and
estimated trajectory is unknown. In addition the system bootstrapping require multi-view or filtering techniques to produce an initial map as it cannot be triangulated from the very 
first frame. Last but not least, monocular SLAM suffers from scale drift and may fail if performing pure rotations in exploration. By using a stereo or 
an RGB-D camera all these issues are solved and allows for the most reliable Visual SLAM solutions.

\begin{figure}[t]
      \centering
      \subfloat[Stereo input: trajectory and sparse reconstruction of an urban environment with multiple loop closures.]
      {
      \includegraphics[width=0.99\linewidth]{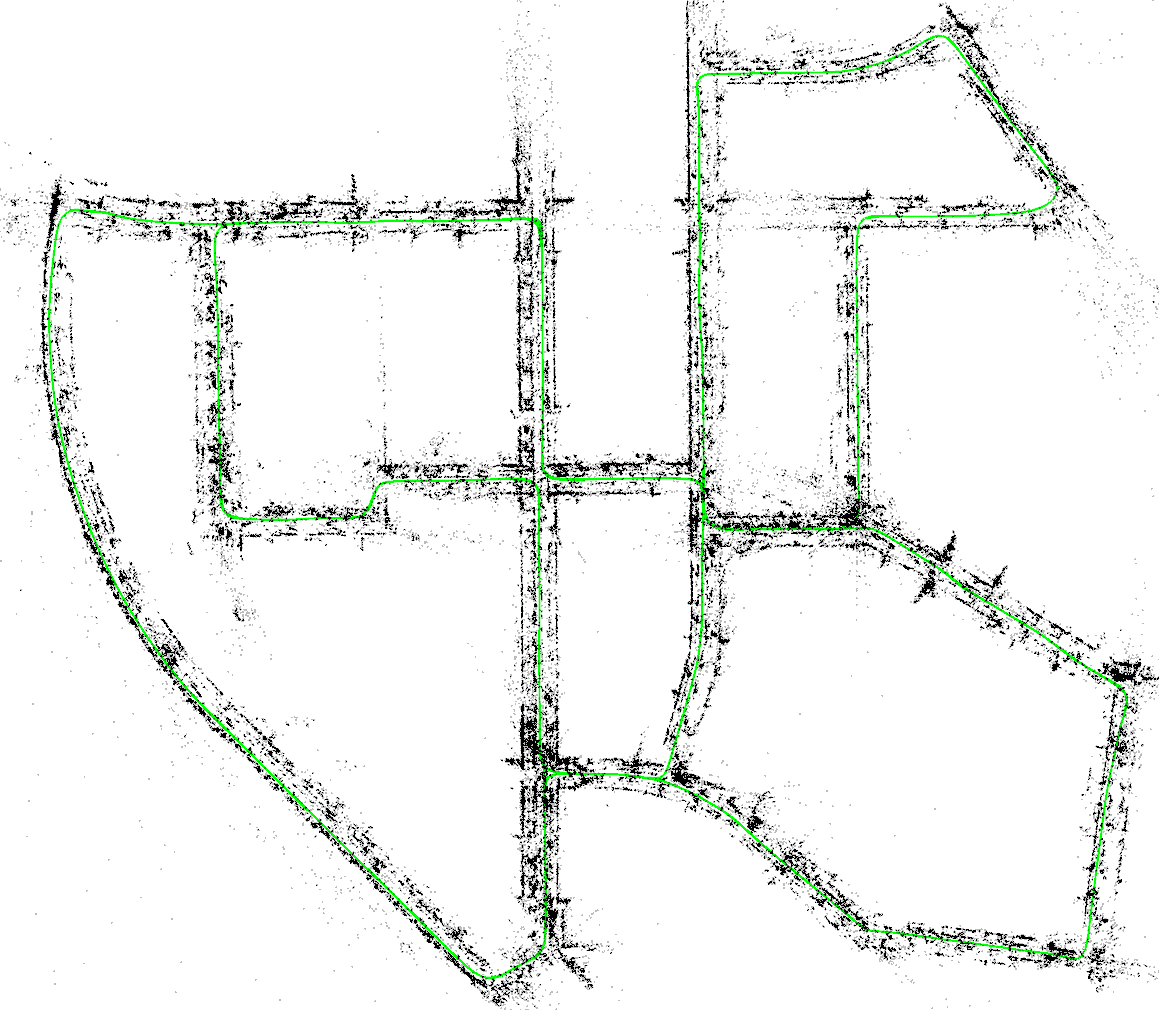}
      }
      
      \subfloat[RGB-D input: keyframes and dense pointcloud of a room scene with one loop closure. The pointcloud is rendered by backprojecting the sensor depth maps 
      from estimated keyframe poses. No fusion is performed.]
      {
      \includegraphics[width=0.99\linewidth]{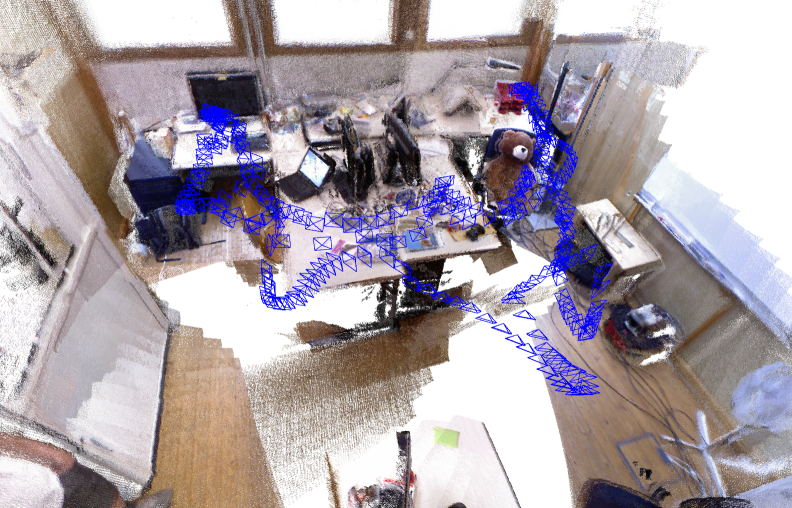}
      }
      
      \caption{ORB-SLAM2 processes stereo and RGB-D inputs to estimate camera trajectory and build a map of the environment. The system is able to close loops, relocalize, and reuse its map in 
      real-time on standard CPUs with high accuracy and robustness.}
      \label{fig:example}
      \end{figure}

In this paper we build on our monocular ORB-SLAM \cite{MurTRO15} and propose ORB-SLAM2 with the following contributions:
\begin{itemize}
 \item The first open-source\footnote{\url{https://github.com/raulmur/ORB_SLAM2}} SLAM system for monocular, stereo and RGB-D cameras, including loop closing, relocalization and map reuse.
 \item Our RGB-D results show that by using Bundle Adjustment (BA) we achieve more accuracy than state-of-the-art methods based on ICP or photometric and depth error minimization.
 \item By using close and far stereo points and monocular observations our stereo results are more accurate than the state-of-the-art direct stereo SLAM.
 \item A lightweight localization mode that can effectively reuse the map with mapping disabled.

\end{itemize}

Fig. \ref{fig:example} shows examples of ORB-SLAM2 output from stereo and RGB-D inputs. The stereo case shows the final trajectory and sparse reconstruction of the sequence 00 from 
the KITTI dataset \cite{KITTI}. This is an urban sequence with multiple loop closures that ORB-SLAM2 was able to successfully detect. The RGB-D case shows the keyframe poses estimated 
in sequence fr1\_room from the TUM RGB-D Dataset \cite{tumrgbd}, and a dense pointcloud, rendered by backprojecting sensor depth maps from the estimated keyframe poses. Note that our SLAM 
does not perform any fusion like KinectFusion \cite{KinectFusion} or similar, but the good 
definition indicates the accuracy of the keyframe poses. More examples are shown on the attached video.

In the rest of the paper, we discuss related work in Section \ref{sec:related}, we describe our system in Section \ref{sec:sys}, then present the evaluation results in Section \ref{sec:eval} and
end with conclusions in Section \ref{sec:conclusion}.

\section{Related Work} \label{sec:related}
In this section we discuss related work on stereo and RGB-D SLAM. Our discussion, as well as the evaluation in Section \ref{sec:eval} is focused only on SLAM approaches. 

\subsection{Stereo SLAM}
A remarkable early stereo SLAM system was the work of Paz et al. \cite{pazTRO08}. Based on Conditionally Independent Divide and Conquer EKF-SLAM it was able
to operate in larger environments than other approaches at that time. Most importantly, it was the first stereo SLAM exploiting both close and far points 
(i.e. points whose depth cannot be reliably estimated due to little disparity in the stereo camera), using an inverse depth parametrization \cite{InverseDepth} for the latter.
They empirically showed that points can be reliably triangulated if their depth is less than ${\sim}40$ times the stereo baseline. In this work we follow this strategy of treating in a different 
way \emph{close} and \emph{far} points, as explained in Section \ref{sec:sec:closefar}. 

Most modern stereo SLAM systems are keyframe-based \cite{WhyFilter} and perform BA optimization in a local area to achieve scalability. The work of Strasdat et al. \cite{DWO} 
performs a joint optimization of BA (point-pose constraints) in an inner window of keyframes and pose-graph (pose-pose constraints) in an outer window. By limiting the 
size of these windows the method achieves constant time complexity, at the expense of not guaranteeing global consistency. The RSLAM of Mei et al. \cite{RSLAM} uses a relative 
representation
of landmarks and poses and performs relative BA in an active area which can be constrained for constant-time. RSLAM is able to close loops which allow to expand 
active areas at both sides of a loop, but global consistency is not enforced. The recent S-PTAM by Pire et al. \cite{sptam} performs local BA, however it 
lacks large loop closing.
Similar to these approaches we perform BA in a local set of keyframes so that the complexity is independent of the map size and we can operate in large environments. 
However our goal is to build a globally consistent map. When closing a loop, our system aligns first both sides, similar to RSLAM, so that the tracking is able to continue localizing 
using the old map and then performs a pose-graph optimization that minimizes the drift accumulated in the loop, followed by full BA. 

The recent Stereo LSD-SLAM of Engel et al. \cite{StereoLSD} is a semi-dense direct approach that minimizes photometric error in image regions with high
gradient. Not relying on features, the method is expected to be more robust to motion blur or poorly-textured environments. However as a direct method its performance
can be severely degraded by unmodeled effects like rolling shutter or non-lambertian reflectance. 

\begin{figure*}[t] 
      \centering
      \subfloat[System Threads and Modules.]
      {
      \includegraphics[width=0.53\linewidth]{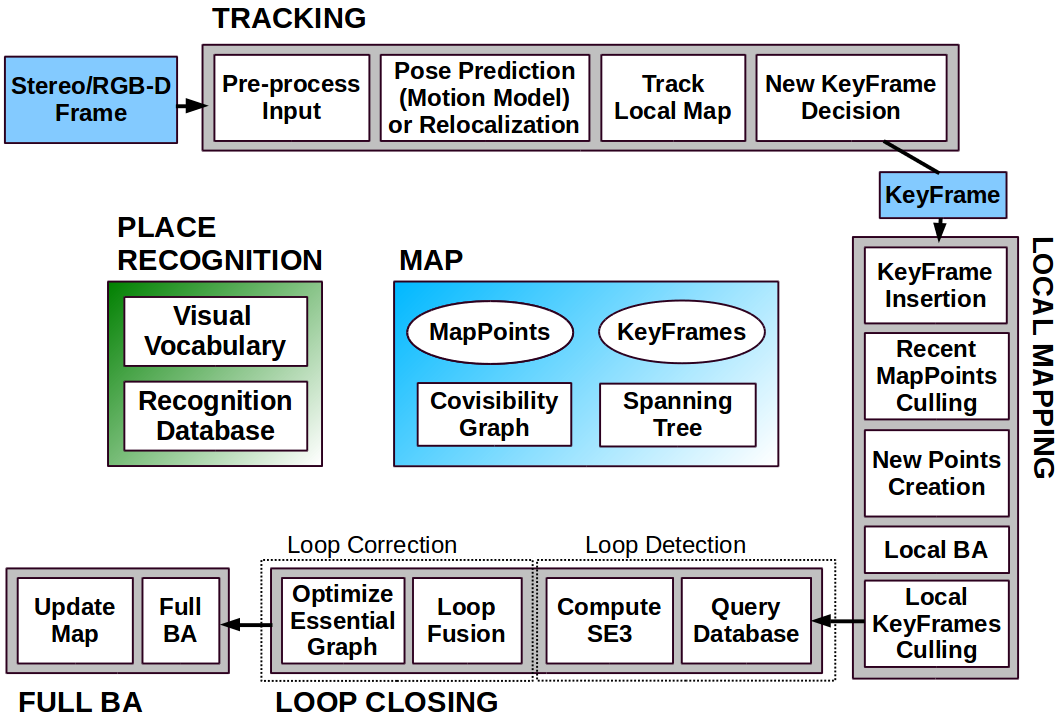}
      }      
      \quad
      \subfloat[Input pre-processing] 
      {
      \label{fig:stereo}
      \includegraphics[width=0.35\linewidth]{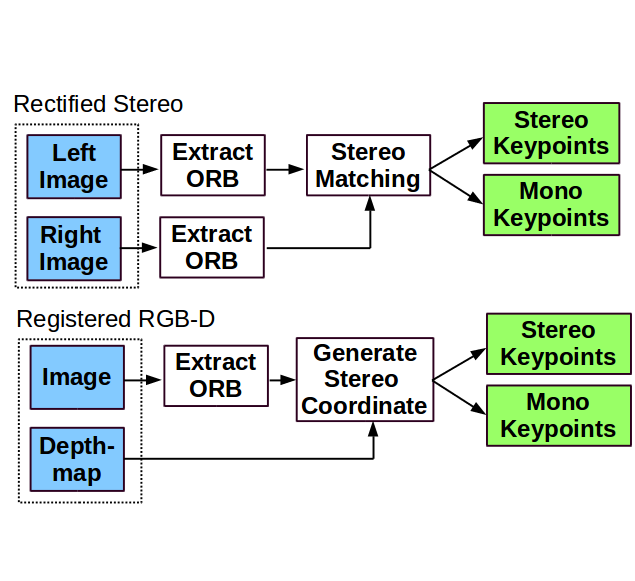}
      }
      \caption{ORB-SLAM2 is composed of three main parallel threads: tracking, local mapping and loop closing, which can create a fourth thread to perform full BA
      after a loop closure. The tracking thread pre-processes the stereo or RGB-D input so that the rest of the system operates independently of the input sensor. Although it is 
      not shown in this figure, ORB-SLAM2 also works with a monocular input as in \cite{MurTRO15}.}
      \label{fig:overview}
   \end{figure*}
\subsection{RGB-D SLAM}

One of the earliest and most famed RGB-D SLAM systems was the KinectFusion of Newcombe et al. \cite{KinectFusion}. This method fused all depth data from the sensor into a volumetric 
dense model that is used to track the camera pose using ICP. This system was limited to small workspaces due to its volumetric representation and the lack of loop closing.
Kintinuous by Whelan et al. \cite{Kintinuous} was able to operate in large environments by using a rolling cyclical buffer and included loop closing using place recognition
and pose graph optimization. 

Probably the first popular open-source system was the RGB-D SLAM of Endres et al. \cite{rgbdslam}. This is a feature-based system, whose front-end computes frame-to-frame motion 
by feature matching and ICP. The back-end performs pose-graph optimization with loop closure constraints from a heuristic search. Similarly the back-end of DVO-SLAM by Kerl et al. \cite{DVOSLAM} 
optimizes a pose-graph where keyframe-to-keyframe constraints are computed from a visual odometry that minimizes both photometric and depth error. DVO-SLAM also searches for 
loop candidates in a heuristic fashion over all previous frames, instead of relying on place recognition.

The recent ElasticFusion of Whelan et al. \cite{ElasticFusion} builds a surfel-based map of the environment. This is a map-centric approach that forget poses and performs 
loop closing applying a non-rigid deformation to the map, instead of a standard pose-graph optimization. The detailed reconstruction and localization accuracy of this system
is impressive, but the current implementation is limited to room-size maps as the complexity scales with the number of surfels in the map.

As proposed by Strasdat et al. \cite{DWO} our ORB-SLAM2 uses depth information to synthesize a stereo coordinate for extracted features on the image. This way our system is
agnostic of the input being stereo or RGB-D. Differently to all above methods our back-end is based on bundle adjustment and builds a globally consistent sparse reconstruction. 
Therefore our method is lightweight and works with standard CPUs. Our goal is long-term and globally consistent localization instead of building the most detailed dense reconstruction.
However from the highly accurate keyframe poses one could fuse depth maps and get accurate reconstruction on-the-fly in a local area or post-process the depth maps from all keyframes
after a full BA and get an accurate 3D model of the whole scene.

\section{ORB-SLAM2} \label{sec:sys}

ORB-SLAM2 for stereo and RGB-D cameras is built on our monocular feature-based ORB-SLAM \cite{MurTRO15}, whose main components are summarized here for reader convenience. 
A general overview of the system is shown in Fig. \ref{fig:overview}. The system has three main parallel threads: 1) the tracking to localize the camera with every frame by finding
feature matches to the local map and minimizing the reprojection error applying motion-only BA, 2) the local mapping to manage the local map and optimize it, performing local BA,
3) the loop closing to detect large loops and correct the accumulated drift by performing a pose-graph optimization. This thread launches a fourth thread to perform full BA after 
the pose-graph optimization, to compute the optimal structure and motion solution.

The system has embedded a Place Recognition module based on DBoW2 \cite{dorian} for relocalization, in case of tracking failure (e.g. an occlusion)  or for reinitialization in an already 
mapped scene, 
and for loop detection. The system maintains a covisibiliy graph \cite{DWO} that links any two keyframes observing common points and a minimum spanning tree connecting all keyframes.
These graph structures allow to retrieve local windows of keyframes, so that tracking and local mapping operate locally, allowing to work on large environments, and serve as 
structure for the pose-graph optimization performed when closing a loop.

The system uses the same ORB features \cite{ORB} for tracking, mapping and place recognition tasks. These features are robust to rotation and scale and present a good invariance 
to camera auto-gain and auto-exposure, and illumination changes. Moreover they are fast to extract and match allowing for real-time operation and show good precision/recall 
performance in bag-of-word place recognition \cite{MurICRA14}.

In the rest of this section we present how stereo/depth information is exploited and which elements of the system are affected. For a detailed description of each system block, we
refer the reader to our monocular publication \cite{MurTRO15}.

\subsection{Monocular, Close Stereo and Far Stereo Keypoints} \label{sec:sec:closefar}
ORB-SLAM2 as a feature-based method pre-processes the input to extract features at salient keypoint locations, as shown in Fig. \ref{fig:stereo}. The input 
images are then discarded and all system operations
are based on these features, so that the system is independent of the sensor being stereo or RGB-D. 
Our system handles monocular and stereo keypoints, which are further classified as close or far.

\textbf{Stereo keypoints} are defined by three coordinates $\mathbf{x}_\mathrm{s} = \left(u_L,v_L,u_R\right)$, being $(u_L,v_L)$ the coordinates on the left image and $u_R$ the horizontal
coordinate in the right image. For stereo cameras, we extract ORB in both images and for every left ORB we search for a match in the right image. This can be done very efficiently assuming
stereo rectified images, so that epipolar lines are horizontal. We then generate the stereo keypoint with the coordinates of the left ORB and the horizontal coordinate of the right
match, which is subpixel refined by patch correlation. For RGB-D cameras, we extract ORB features on 
the RGB image and, as proposed by Strasdat et al. \cite{DWO}, for each feature with coordinates 
$\left(u_L,v_L\right)$ we transform its depth value $d$ into a virtual right coordinate:
\begin{equation}
 u_R = u_L - \frac{f_x b}{d}
\end{equation}
where $f_x$ is the horizontal focal length and  $b$ is the baseline between the 
structured light projector and the infrared camera, which we approximate to 8cm for Kinect 
and Asus Xtion. The uncertainty of the depth sensor is represented by the uncertainty of the 
virtual right coordinate. In this way, features from stereo and RGB-D input are handled equally 
by the rest of the system.

A stereo keypoint is classified as \textbf{close} if its associated depth is less than 40 times the stereo/RGB-D baseline, as suggested in \cite{pazTRO08}, otherwise it is 
classified as \textbf{far}. Close keypoints can be safely triangulated from one frame as depth is accurately estimated and provide scale, translation and rotation information.
On the other hand far points provide accurate rotation information but weaker scale and translation information. We triangulate far points when they are supported by multiple views.

\textbf{Monocular keypoints} are defined by two coordinates $\mathbf{x}_\mathrm{m} = \left(u_L,v_L\right)$ on the left image and correspond to all those ORB for which a stereo match 
could not be found or that have an invalid depth value in the RGB-D case. These points are only triangulated from multiple views and do not provide scale information, but contribute
to the rotation and translation estimation.

\subsection{System Bootstrapping}
One of the main benefits of using stereo or RGB-D cameras is that, by having depth information from just one frame, we do not need a specific structure from motion initialization 
as in the monocular case. At system startup we create a keyframe with the first frame, set its pose to the origin, and create an initial map from all stereo keypoints.
\subsection{Bundle Adjustment with Monocular and Stereo Constraints}

Our system performs BA to optimize the camera pose in the tracking thread (motion-only BA), to optimize a local window of keyframes and points in the local mapping thread (local BA),
and after a loop closure to optimize all keyframes and points (full BA).  We use the Levenberg--Marquardt method implemented in g2o \cite{g2o}.

\textbf{Motion-only BA} optimizes the camera orientation $\mathbf{R}\in SO(3)$ and position $\mathbf{t}\in \mathbb{R}^3$, minimizing the reprojection
error between matched 3D points $\mathbf{X}^i \in\mathbb{R}^3$ in world coordinates
and keypoints $\mathbf{x}^i_\mathrm{(\cdot)}$, either monocular $\mathbf{x}^i_\mathrm{m}\in\mathbb{R}^2$ or stereo  $\mathbf{x}^i_\mathrm{s}\in\mathbb{R}^3$, with $i\in\mathcal{X}$ 
the set of all matches:

\begin{equation}
 \{\mathbf{R}, \mathbf{t}\} = \argmin_{\mathbf{R}, \mathbf{t}}  
 \sum_{i\in\mathcal{X}} \rho\left(\left\|\mathbf{x}^i_\mathrm{(\cdot)}-\pi_\mathrm{(\cdot)}\left(\mathbf{R}\mathbf{X}^i + \mathbf{t}\right)\right\|^2_\Sigma\right)
 \end{equation}
 where $\rho$ is the robust Huber cost function and $\Sigma$ the covariance matrix associated to the scale of the keypoint. The projection functions $\pi_{(\cdot)}$,  
 monocular $\pi_\mathrm{m}$ and rectified stereo $\pi_\mathrm{s}$,
 are defined as follows:
\begin{equation}
 \pi_\mathrm{m} \left(\begin{bmatrix}X\\Y\\Z\end{bmatrix} \right) = \begin{bmatrix} f_x\frac{X}{Z} + c_x\\[0.2em] f_y\frac{Y}{Z} + c_y \end{bmatrix}
,
\pi_\mathrm{s} \left(\begin{bmatrix}X\\Y\\Z\end{bmatrix} \right) = \begin{bmatrix} f_x\frac{X}{Z} + c_x\\[0.2em] f_y\frac{Y}{Z} + c_y \\[0.2em] f_x\frac{X-b}{Z} + c_x\end{bmatrix}
 \end{equation}
where $(f_x,f_y)$ is the focal length, $(c_x,c_y)$ is the principal point and $b$ the baseline, all known from calibration.

\textbf{Local BA} optimizes a set of covisible keyframes $\mathcal{K}_L$ and all points seen in those keyframes $\mathcal{P}_L$. 
All other keyframes $\mathcal{K}_F$, not in $\mathcal{K}_L$, observing points in $\mathcal{P}_L$
 contribute to the cost function but remain fixed in the optimization. Defining $\mathcal{X}_k$ as the set of matches between points
 in $\mathcal{P}_L$ and keypoints in a keyframe $k$, the optimization problem is the following:
\begin{equation}
\begin{gathered}
 \{\mathbf{X}^i, \mathbf{R}_l, \mathbf{t}_l |  i\in\mathcal{P}_L,  l\in\mathcal{K}_L\} = \argmin_{\mathbf{X}^i, \mathbf{R}_l, \mathbf{t}_l}   
 \sum_{k\in \mathcal{K}_L\cup\mathcal{K}_F} \sum_{j\in\mathcal{X}_k} \rho\left(E_{kj}\right)
 \\
 E_{kj} = \left\|\mathbf{x}^j_\mathrm{(\cdot)}-\pi_\mathrm{(\cdot)}\left(\mathbf{R}_k\mathbf{X}^j + \mathbf{t}_k\right)\right\|^2_\Sigma
 \end{gathered}
 \end{equation}
 
 \textbf{Full BA} is the specific case of local BA, where all keyframes and points in the map are optimized, except the origin keyframe that is fixed to eliminate the gauge freedom.
 
\subsection{Loop Closing and Full BA}

Loop closing is performed in two steps, firstly a loop has to be detected and validated, and secondly the loop is corrected optimizing a pose-graph. 
In contrast to monocular ORB-SLAM, where scale drift may occur \cite{HaukeScale}, the stereo/depth information makes scale observable and the geometric validation and 
pose-graph optimization no longer require dealing with scale drift and are based on rigid body transformations instead of similarities.

In ORB-SLAM2 we have incorporated a full BA optimization after the pose-graph to achieve the optimal solution. This optimization might be very costly and therefore we 
perform it in a separate thread, allowing the system to continue creating map and detecting loops. However this brings the challenge of merging the bundle adjustment
output with the current state of the map. If a new loop is detected while the optimization is running, we abort the optimization
and proceed to close the loop, which will launch the full BA optimization again. When the full BA finishes, 
we need to merge the updated subset of keyframes and points optimized by the full BA, with the non-updated keyframes and points
that where inserted while the optimization was running. This is done by  propagating the correction of updated keyframes (i.e. the transformation from the non-optimized to the optimized pose) 
to non-updated keyframes through the spanning tree. Non-updated points are transformed according to the correction applied to their reference keyframe.

\subsection{Keyframe Insertion}

\begin{figure}[t]
      \centering
      \includegraphics[width=\linewidth]{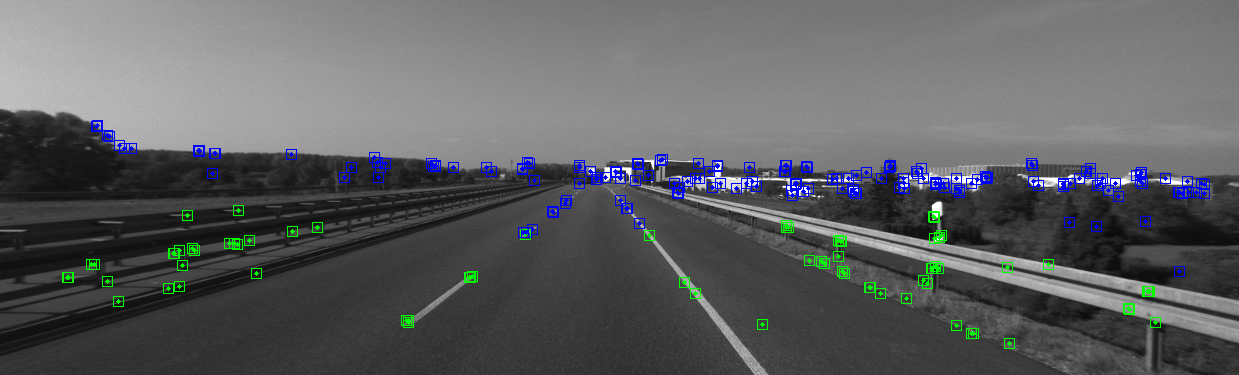}
      \caption{Tracked points in KITTI 01 \cite{KITTI}. Green points have a depth less than 40 times the stereo baseline, while blue points are further away. In this kind of sequences
      it is important to insert keyframes often enough so that the amount of close points allows for accurate translation estimation. Far points contribute to estimate orientation but provide
      weak information for translation and scale.}
      \label{fig:highway}
   \end{figure}

ORB-SLAM2 follows the policy introduced in monocular ORB-SLAM of inserting keyframes very often and culling redundant ones afterwards. The distinction between close and far stereo
points allows us to introduce a new condition for keyframe insertion, which can be critical in challenging environments where a big part of the scene is far from the stereo sensor,
as shown in Fig. \ref{fig:highway}. In such environment we need to have a sufficient amount of close points to accurately estimate translation, therefore if the number of tracked 
close points drops below $\tau_t$ and the frame could create at least $\tau_c$ new close stereo points, the system will insert a new keyframe. We empirically
found that $\tau_t=100$ and $\tau_c=70$ works well in all our experiments.

\subsection{Localization Mode}

We incorporate a Localization Mode which can be useful for lightweight long-term localization in well mapped areas, as long as there are not significant changes in the environment. In this mode the local mapping and loop closing threads are deactivated
and the camera is continuously localized by the tracking using relocalization if needed. In this mode the tracking leverages visual odometry matches and matches to map points.
Visual odometry matches are matches between ORB in the current frame and 3D points created in the previous frame from the stereo/depth information. These matches make the localization
robust to unmapped regions, but drift can be accumulated. Map point matches ensure drift-free localization to the existing map.
This mode is demonstrated in the accompanying video.

\section{Evaluation} \label{sec:eval}

We have evaluated ORB-SLAM2 in three popular datasets and compared to other state-of-the-art SLAM systems, using always the results published by the original authors and standard evaluation metrics in the literature.
We have run ORB-SLAM2 in an Intel Core i7-4790 desktop computer with 16Gb RAM. In order to account for the non-deterministic nature of the multi-threading system, 
we run each sequence 5 times and show median results for the accuracy of the estimated trajectory.
Our open-source implementation includes the calibration and instructions to run the system in all these datasets.

\subsection{KITTI Dataset}

The KITTI dataset \cite{KITTI} contains stereo sequences recorded from a car in urban and highway environments. The stereo sensor has a ${\sim}54 \textrm{cm}$ baseline and works at 10Hz with a resolution
after rectification of $1240 \times 376$ pixels. Sequences 00, 02, 05, 06, 07 and 09 contain loops. Our ORB-SLAM2 detects all loops and is able to reuse its map afterwards, except for sequence 09 
where the loop happens in very few frames at the end of the sequence. Table \ref{tb:kitti} shows results in the 11 training sequences, which have public ground-truth, compared to the state-of-the-art 
Stereo LSD-SLAM \cite{StereoLSD}, to our knowledge the only stereo SLAM showing detailed results for all sequences.
We use two different metrics, the absolute translation RMSE $t_{abs}$ proposed in \cite{tumrgbd}, and the average relative translation $t_{rel}$ and rotation $r_{rel}$ errors proposed in \cite{KITTI}.
Our system outperforms Stereo LSD-SLAM in most sequences, and achieves in general a relative error lower than 1\%. The sequence 01, see Fig. \ref{fig:highway}, is the only highway sequence in the 
training set and the translation error is slightly worse. Translation is harder to estimate in this sequence because very few close points can be tracked, due to
high speed and low frame-rate. However orientation can be accurately estimated, achieving an error of 0.21 degrees per 100 meters, as there are many far point that can be long tracked.
Fig. \ref{fig:kitti} shows some examples of estimated trajectories.

Compared to the monocular results presented in \cite{MurTRO15}, the proposed stereo version is able to process the sequence 01 where the monocular system failed.
In this highway sequence, see Fig. \ref{fig:highway}, close points are in view only for a few frames. The ability of the stereo version to create points from just one stereo keyframe
instead of the delayed initialization of the monocular, consisting on finding matches between two keyframes, is critical in this sequence not to lose tracking.
Moreover the stereo system estimates the map and trajectory with metric scale and does not suffer from scale drift, as seen in Fig. \ref{fig:drift}.

   \begin{table}[t] 
\caption{Comparison of accuracy in the KITTI Dataset.}
\label{tb:kitti}
\begin{center}
\begin{tabular}{|c|c|c|c|c|c|c|}
\hline
         & \multicolumn{3}{|c|}{ORB-SLAM2 (stereo)}  &  \multicolumn{3}{|c|}{Stereo LSD-SLAM} \\
\hline
 \multirow{2}{*}{Sequence} & $t_{rel}$ & $r_{rel}$ & $t_{abs}$  & $t_{rel}$  & $r_{rel}$ & $t_{abs}$  \\
  & (\%) &  (deg/100m) & (m) &  (\%) &  (deg/100m) & (m) \\
 \hline
  \hline
 00 & 0.70 & \textbf{0.25}  & 1.3 & \textbf{0.63} & 0.26 & \textbf{1.0} \\
 01 & \textbf{1.39} & \textbf{0.21} & 10.4 & 2.36 & 0.36 & \textbf{9.0} \\
 02 & \textbf{0.76} & 0.23 & 5.7 & 0.79 & 0.23 & \textbf{2.6} \\
 03 & \textbf{0.71} & \textbf{0.18} & \textbf{0.6} & 1.01 & 0.28 & 1.2 \\
 04 & 0.48 & \textbf{0.13} & 0.2 & \textbf{0.38} & 0.31 & 0.2 \\
 05 & \textbf{0.40} & \textbf{0.16} & \textbf{0.8} & 0.64 & 0.18& 1.5 \\
 06 & \textbf{0.51} & \textbf{0.15} & \textbf{0.8} &  0.71 & 0.18& 1.3 \\
 07 & \textbf{0.50} & \textbf{0.28} & 0.5 & 0.56 & 0.29& 0.5 \\
 08 & \textbf{1.05} & 0.32 & \textbf{3.6} & 1.11 & \textbf{0.31}& 3.9\\
 09 & \textbf{0.87} & 0.27 & \textbf{3.2} & 1.14 & \textbf{0.25}& 5.6\\
 10 & \textbf{0.60} & \textbf{0.27}& \textbf{1.0}& 0.72 & 0.33& 1.5\\
\hline
\end{tabular}
\end{center}
\end{table}

\newcommand{\scale}{0.49}
\begin{figure}[t]
      \centering
      \includegraphics[width=\scale\linewidth]{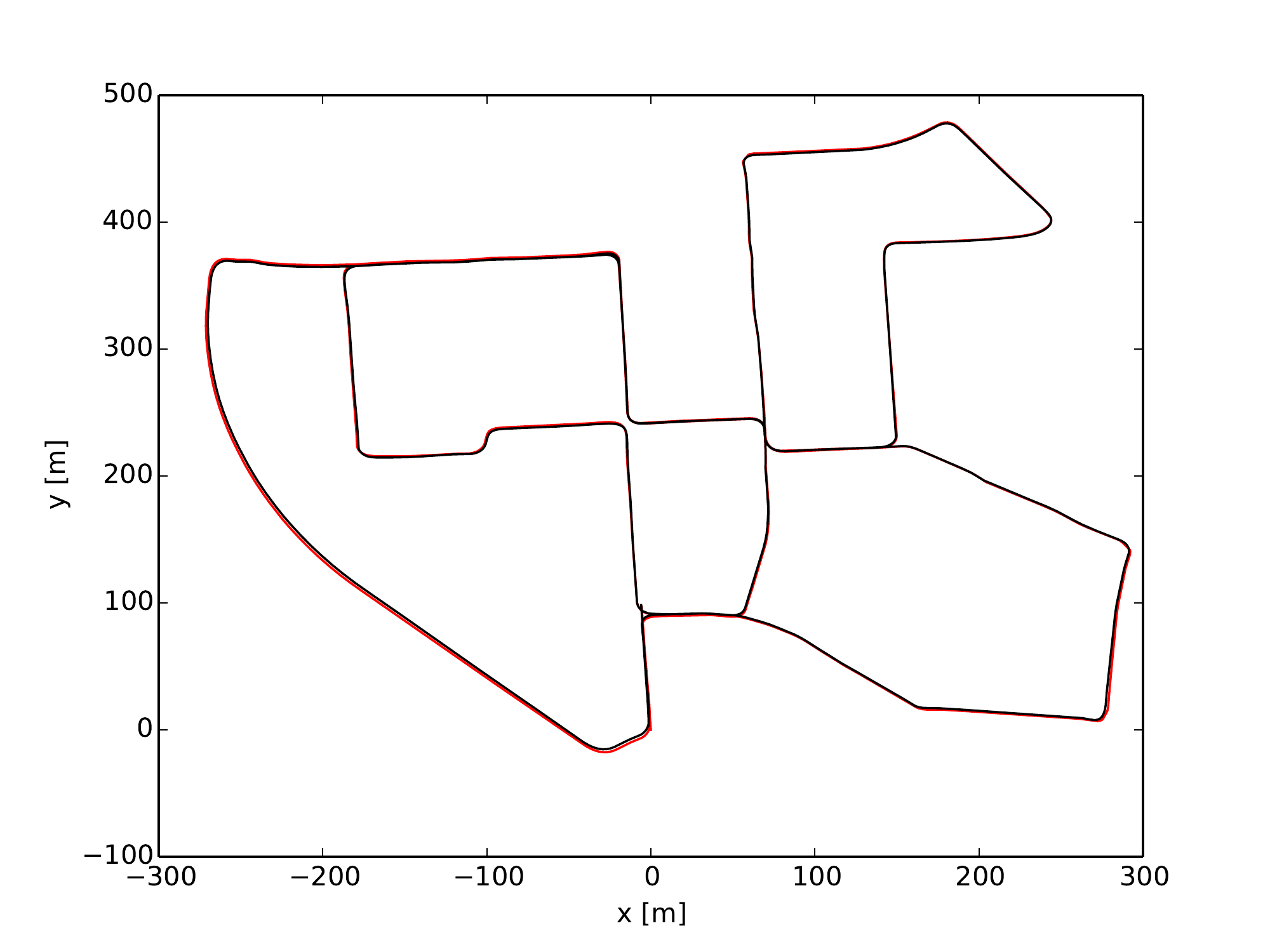}
      \includegraphics[width=\scale\linewidth]{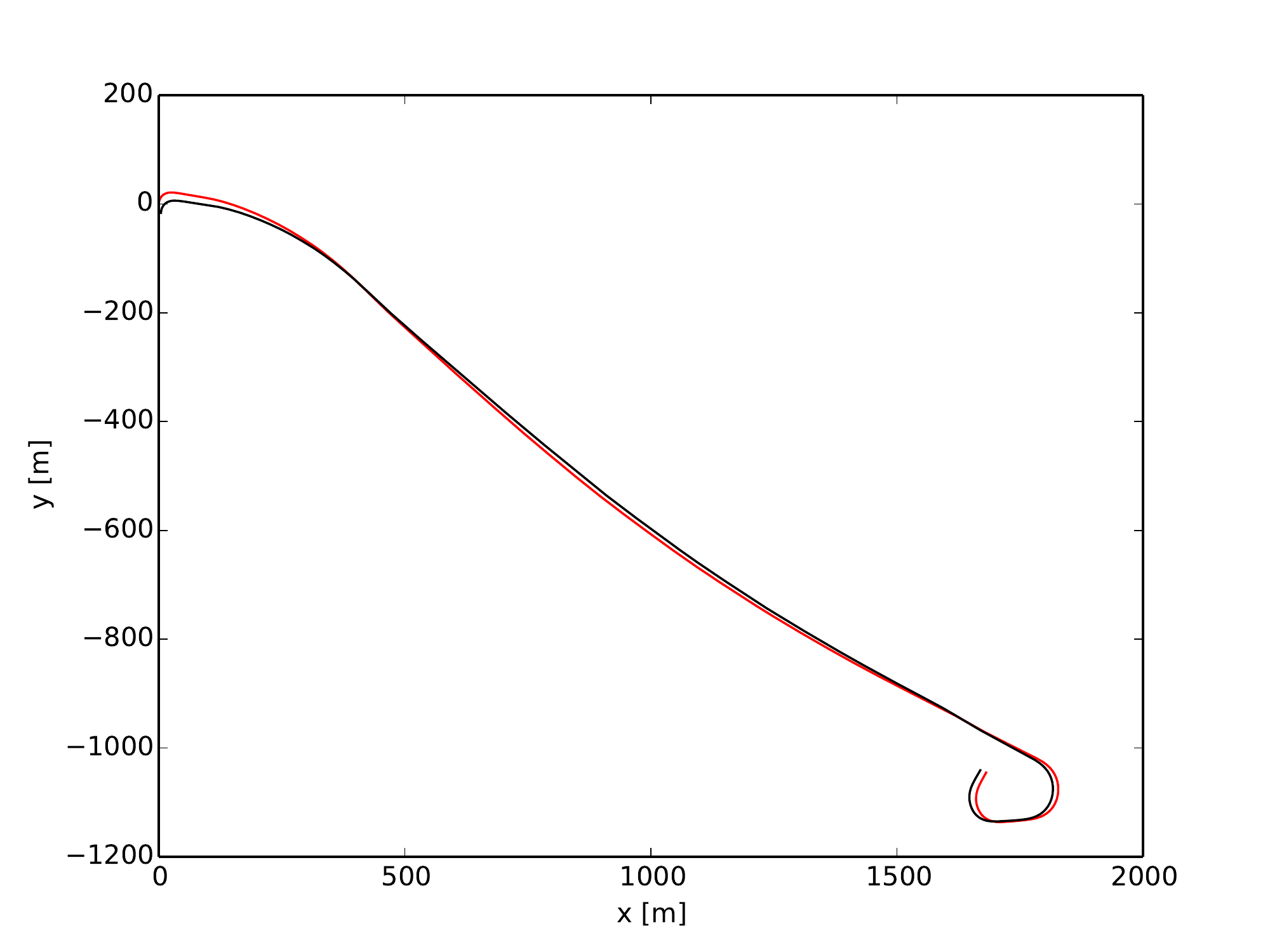}
           
      \includegraphics[width=\scale\linewidth]{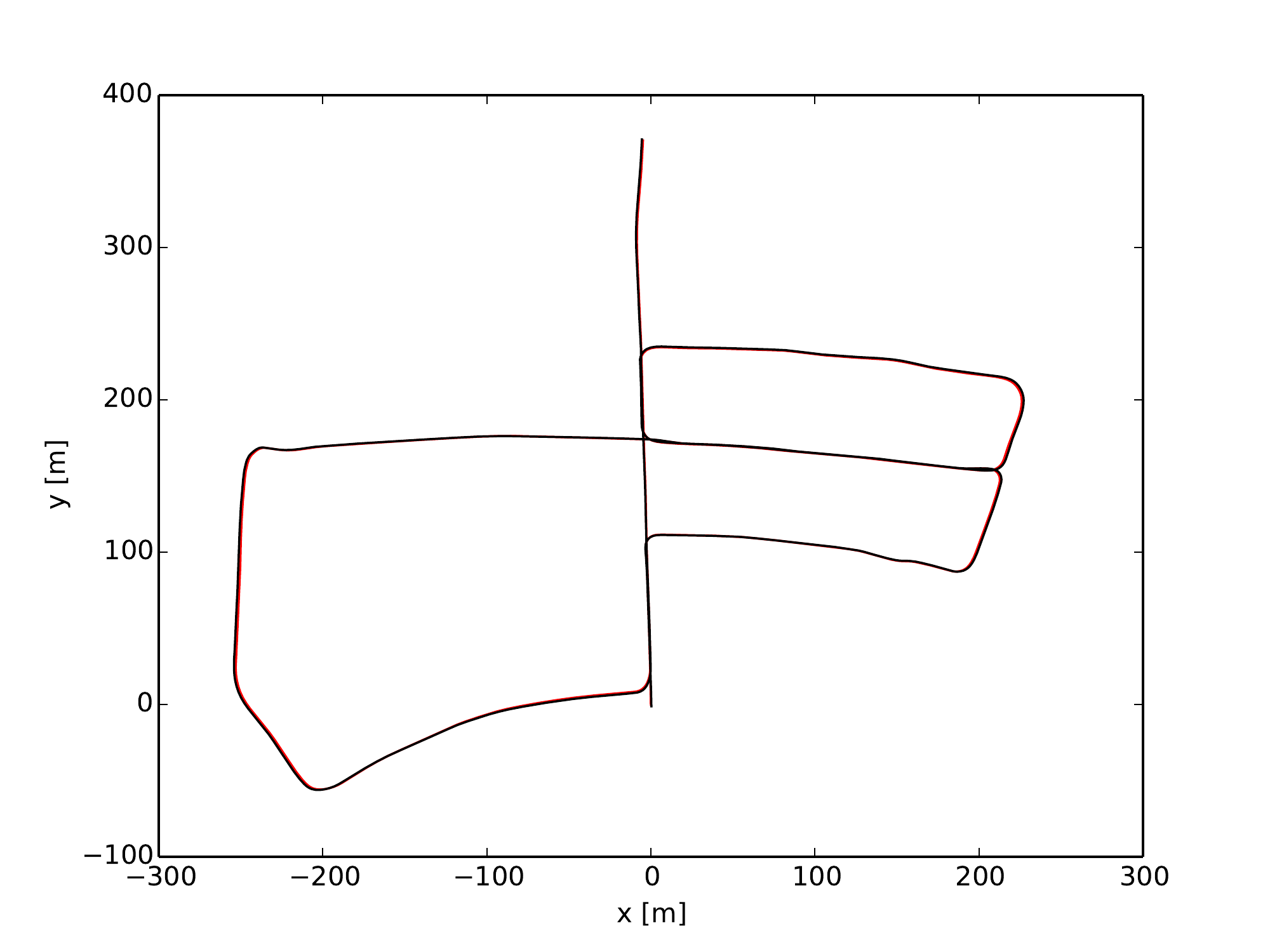}
      \includegraphics[width=\scale\linewidth]{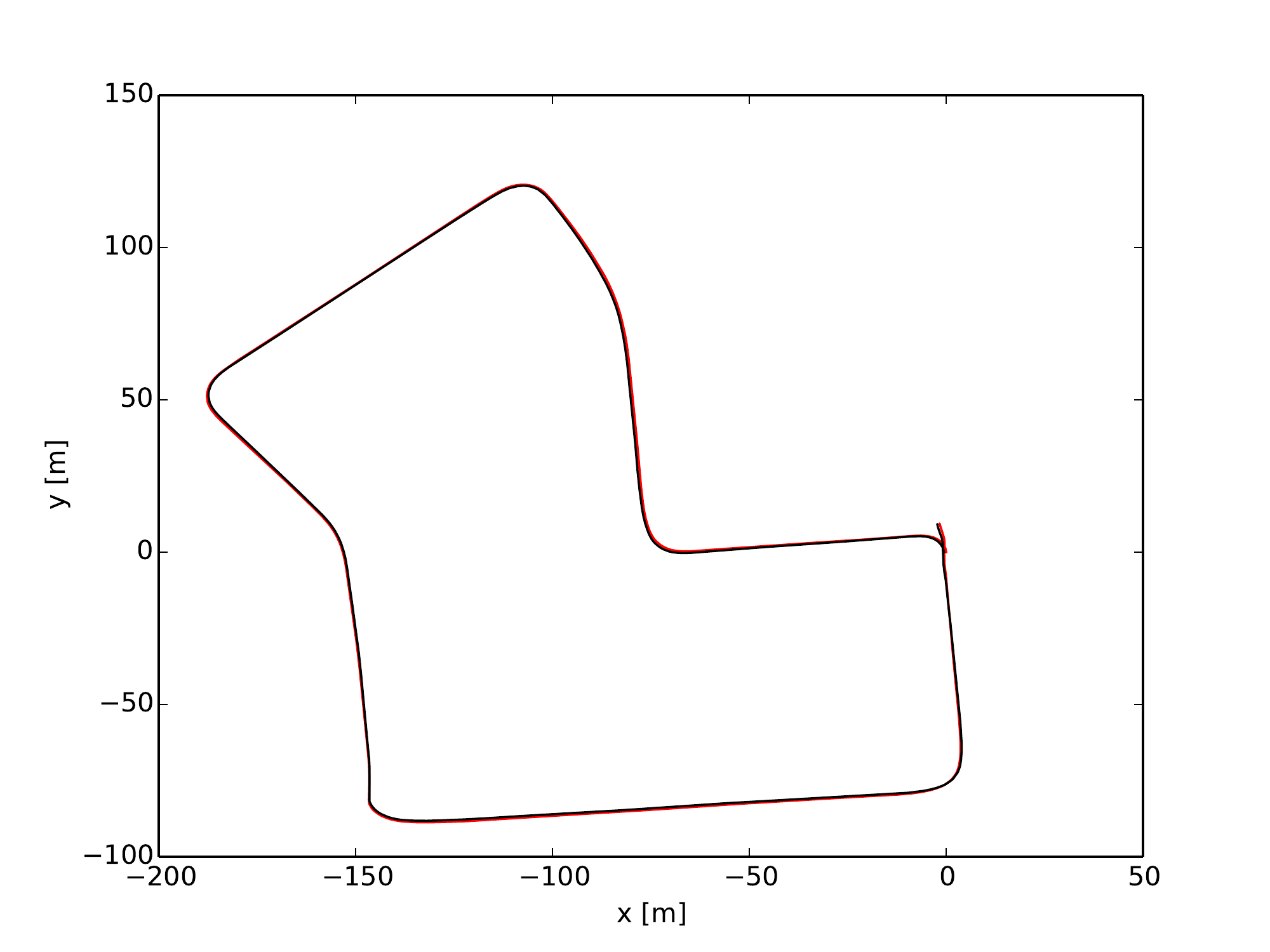} 
      
      \caption{Estimated trajectory (black) and ground-truth (red) in KITTI 00, 01, 05 and 07.}
      \label{fig:kitti}
   \end{figure}
   
\begin{figure}[t]
      \centering
      \includegraphics[width=\scale\linewidth]{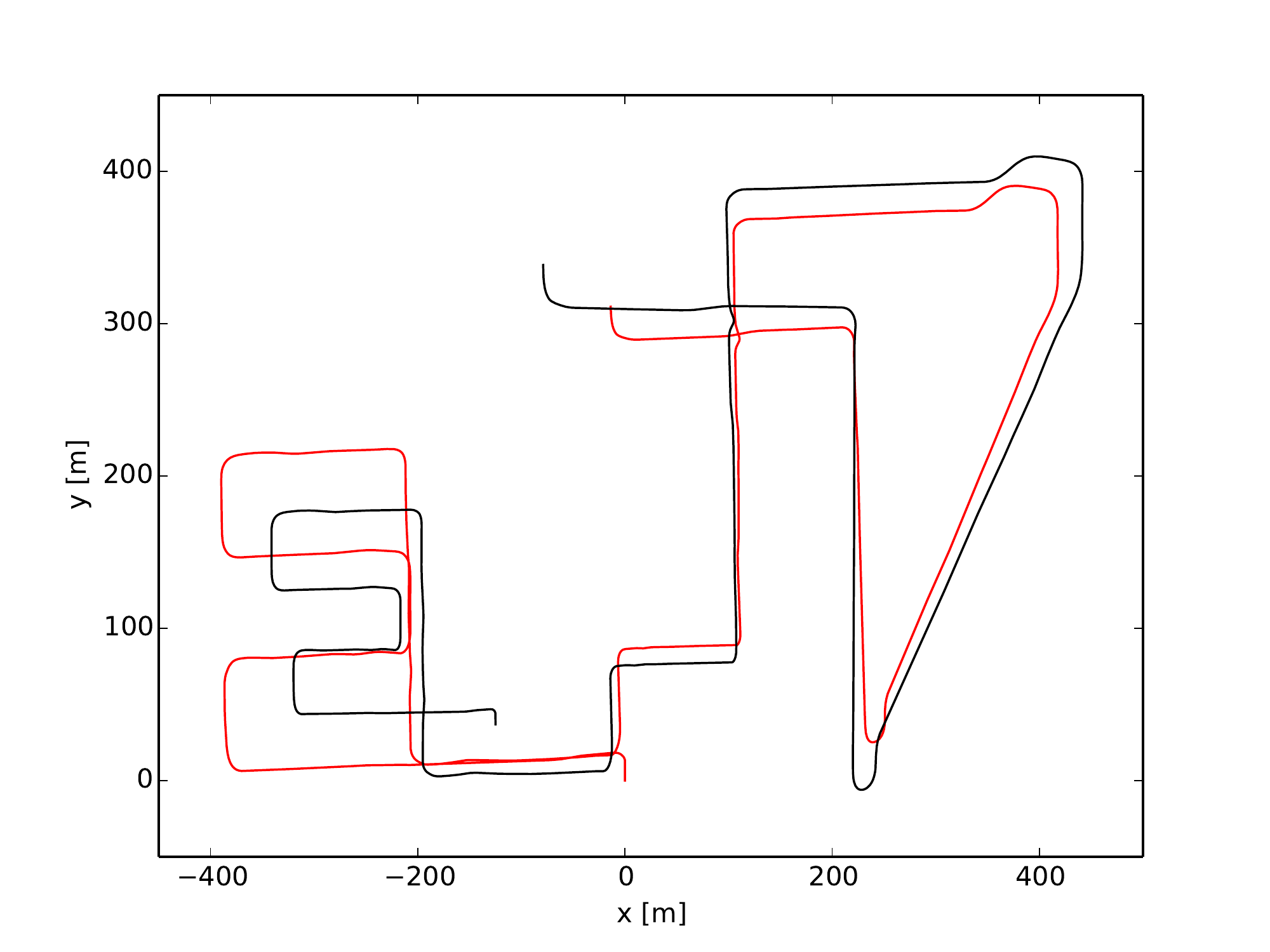}
      \includegraphics[width=\scale\linewidth]{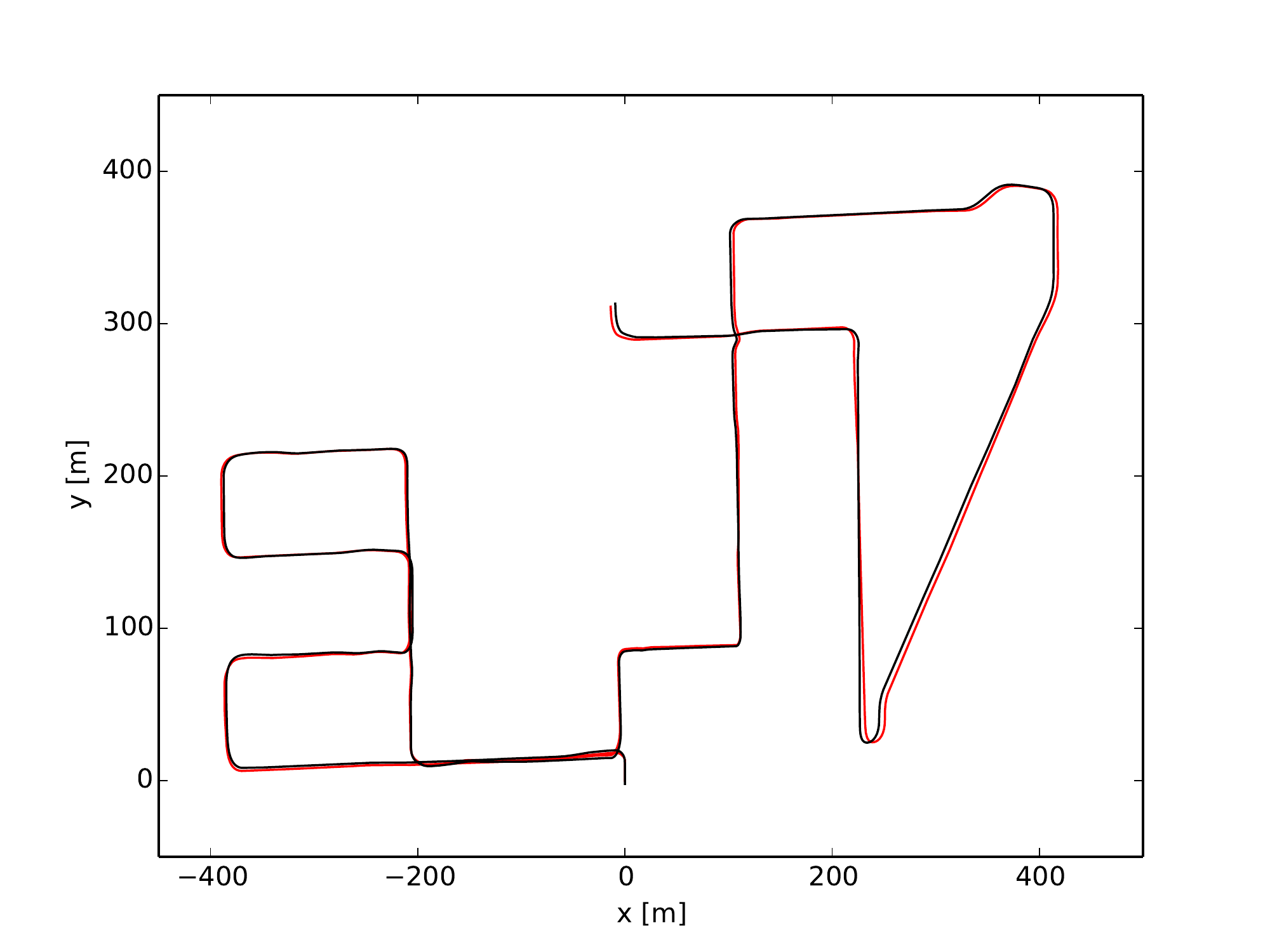}
      
      \caption{Estimated trajectory (black) and ground-truth (red) in KITTI 08. Left: monocular ORB-SLAM \cite{MurTRO15}, right: ORB-SLAM2 (stereo).
      Monocular ORB-SLAM suffers from severe scale drift in this sequence, especially at the turns. In contrast the proposed stereo version is able to estimate
      the true scale of the trajectory and map without scale drift.}
      \label{fig:drift}
   \end{figure}

\subsection{EuRoC Dataset}

\begin{figure}[t]
      \centering
      \includegraphics[width=\scale\linewidth]{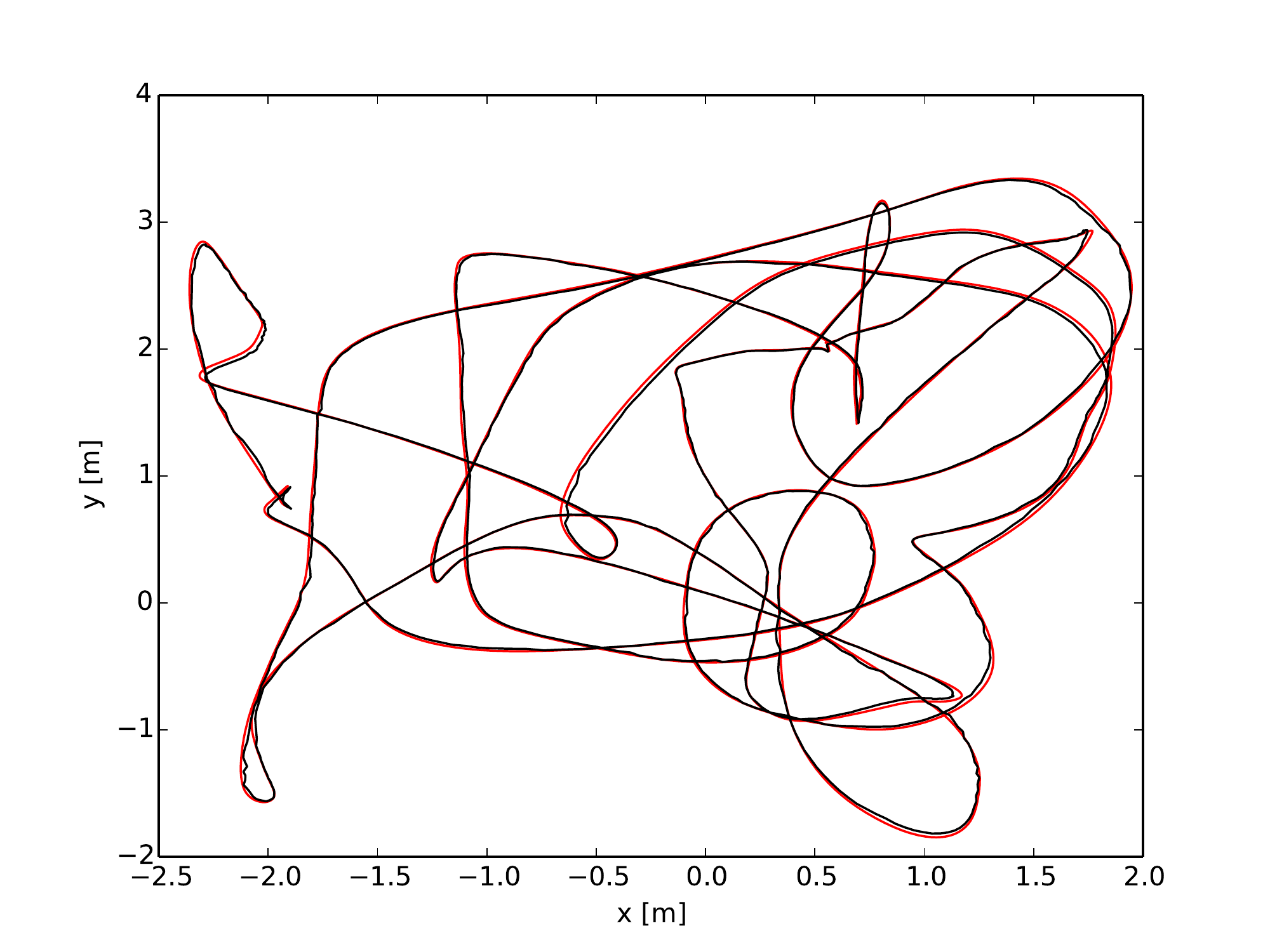}
      \includegraphics[width=\scale\linewidth]{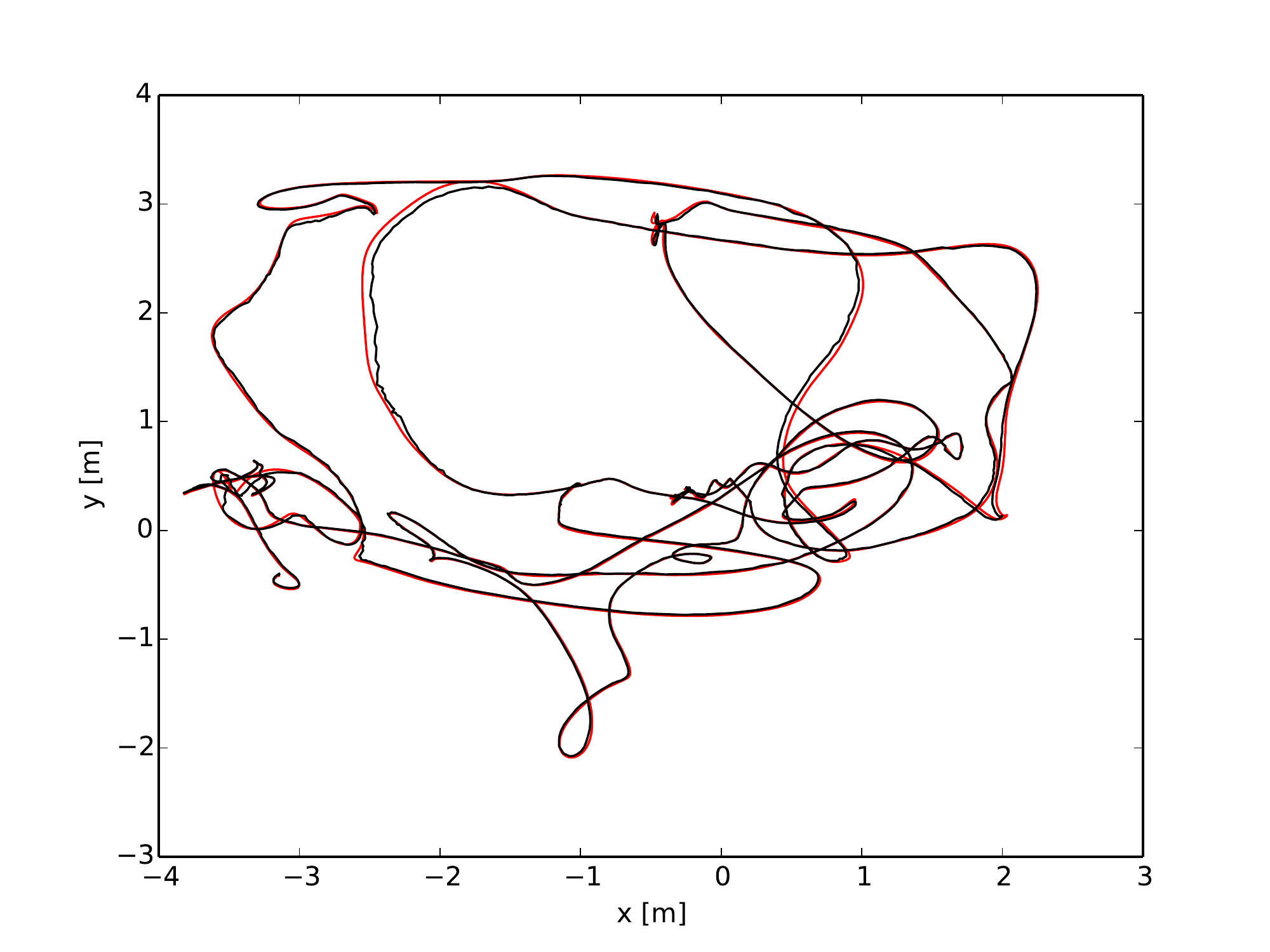}
           
      \includegraphics[width=\scale\linewidth]{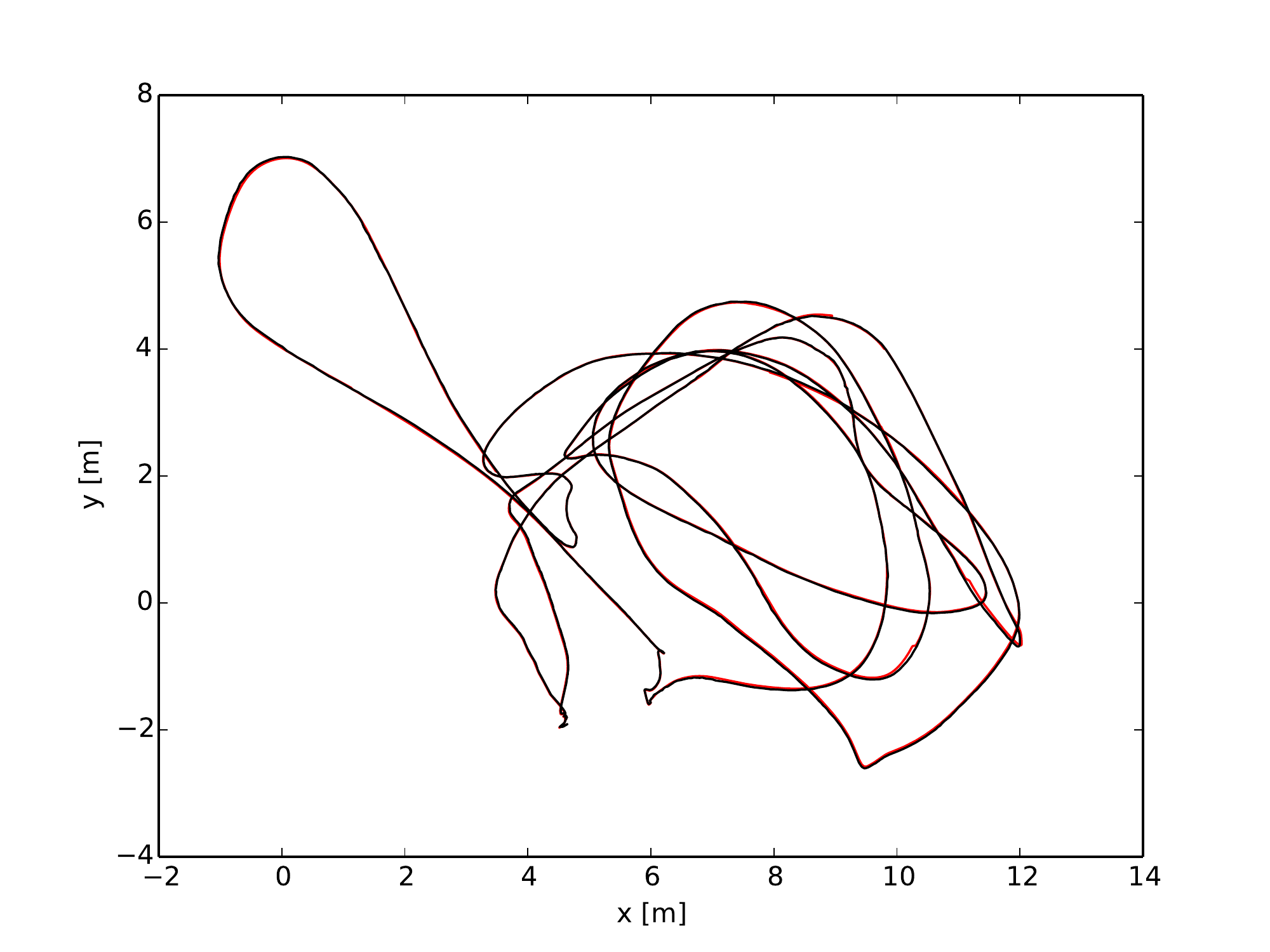}
      \includegraphics[width=\scale\linewidth]{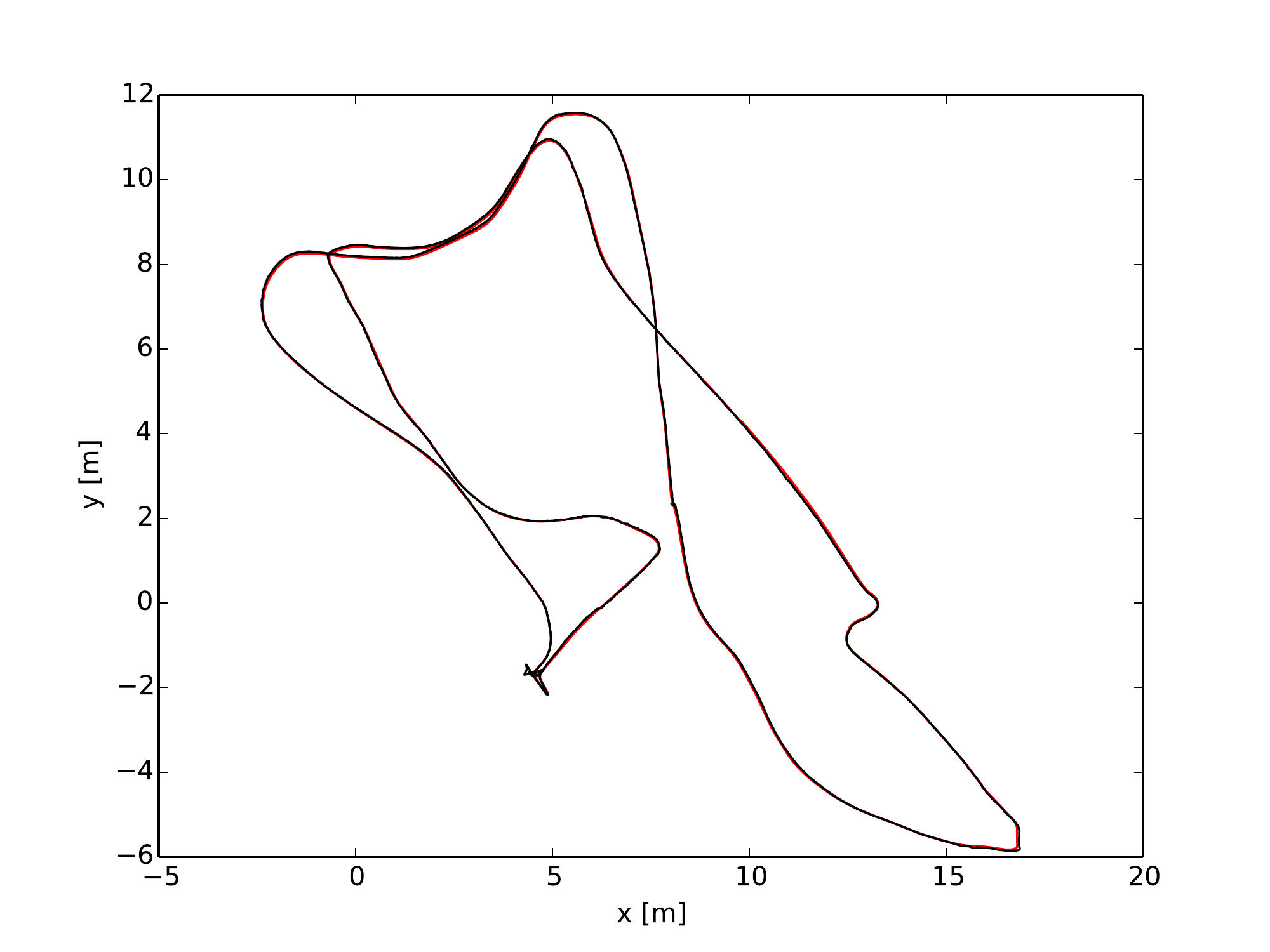}  
      \caption{Estimated trajectory (black) and groundtruth (red) in EuRoC 
      V1\_02\_medium, V2\_02\_medium, MH\_03\_medium and MH\_05\_difficutlt.}
      \label{fig:euroc}
   \end{figure}

The recent EuRoC dataset \cite{euroc} contains 11 stereo sequences recorded from a micro aerial vehicle (MAV) flying around two different rooms and a large industrial environment. 
The stereo sensor has a ${{\sim}11}\textrm{cm}$ baseline and provides WVGA images at 20Hz. The sequences are classified as \emph{easy} , \emph{medium} and \emph{difficult} depending on 
MAV's speed, illumination and scene texture. In all sequences the MAV revisits the environment and ORB-SLAM2 is able to reuse its map, closing loops when necessary. 
Table \ref{tb:euroc} shows absolute translation RMSE of ORB-SLAM2 for all sequences, comparing to Stereo LSD-SLAM, for the results provided in \cite{StereoLSD}.
ORB-SLAM2 achieves a localization precision of a few centimeters and is more accurate than Stereo LSD-SLAM. Our tracking get lost in some parts of 
\emph{V2\_03\_difficult} due to severe motion blur. As shown in \cite{ORBIMU}, this sequence can be processed using IMU information. Fig. \ref{fig:euroc} shows examples of computed trajectories compared to the ground-truth.

\subsection{TUM RGB-D Dataset}

\begin{figure*}[t]
      \centering
      \includegraphics[width=\linewidth]{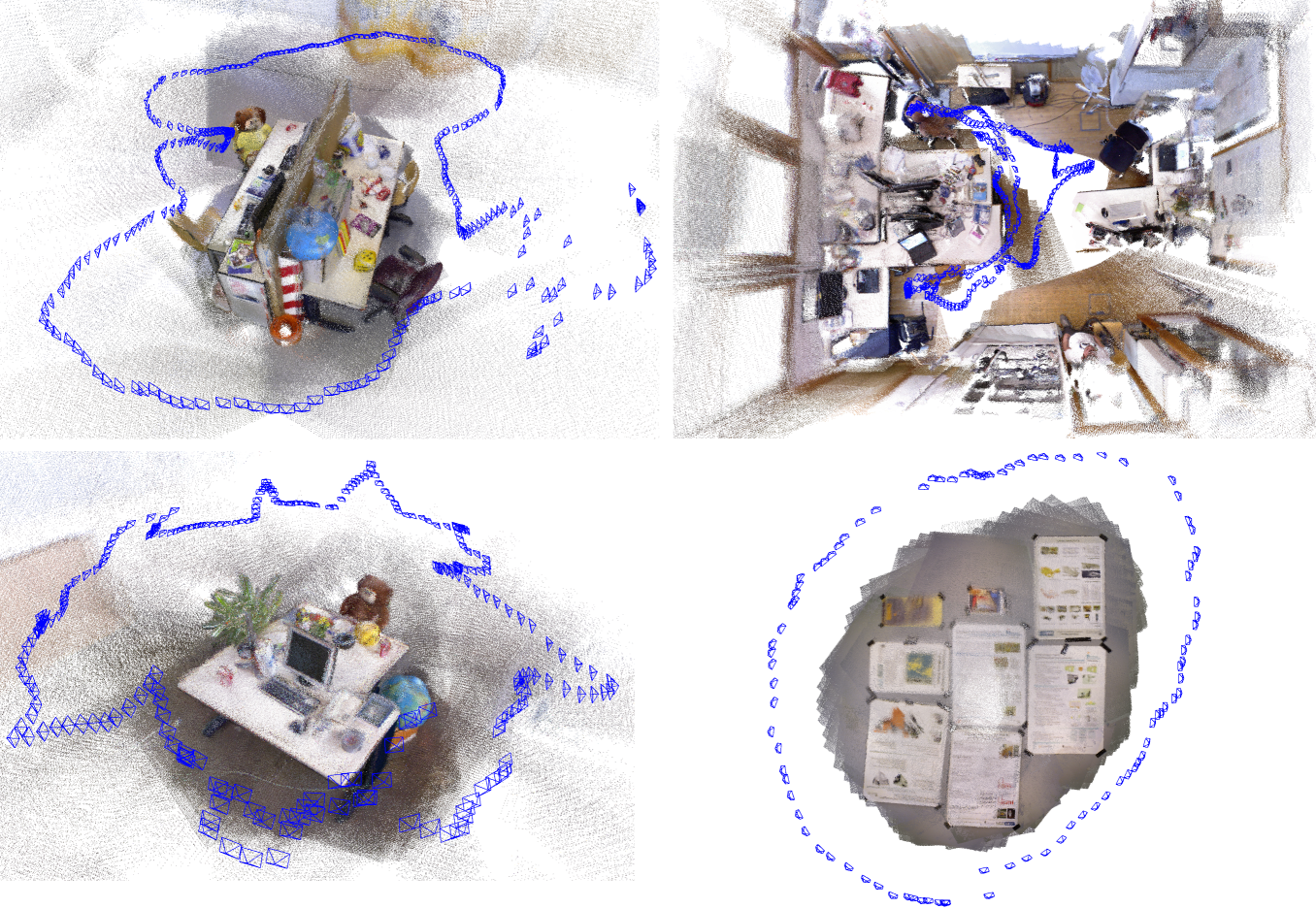}
      \caption{Dense pointcloud reconstructions from estimated keyframe poses and sensor depth maps in TUM RGB-D \emph{fr3\_office}, \emph{fr1\_room}, \emph{fr2\_desk} and \emph{fr3\_nst}.}
      \label{fig:reconstructions}
   \end{figure*}

The TUM RGB-D dataset \cite{tumrgbd} contains indoors sequences from RGB-D sensors grouped in several categories to evaluate object reconstruction and SLAM/odometry methods under
different texture, illumination and structure conditions. We show results in a subset of sequences where most RGB-D methods are usually evaluated. In Table \ref{tb:tum} we compare our accuracy to the following state-of-the-art methods: 
ElasticFusion \cite{ElasticFusion}, Kintinuous \cite{Kintinuous}, DVO-SLAM \cite{DVOSLAM} and RGB-D SLAM \cite{rgbdslam}. Our method is the only one based on bundle adjustment and outperforms 
the other approaches in most sequences. As we already noticed for RGB-D SLAM results in \cite{MurTRO15}, depthmaps for \emph{freiburg2} sequences have a 4$\%$ scale bias, probably coming 
from miscalibration, that we have compensated in our runs and could partly explain our significantly better results.
Fig. \ref{fig:reconstructions} shows the point clouds that result from backprojecting the sensor depth maps from the computed keyframe poses 
in four sequences. 
The good definition and the straight contours of desks and posters prove the high accuracy localization of our approach.

\begin{table}[t] 
\caption{EuRoC Dataset. Comparison of Translation RMSE (${m}$).}
\label{tb:euroc}
\begin{center}
\begin{tabular}{|l|c|c|}
\hline
 Sequence      &  ORB-SLAM2 (stereo)  & Stereo LSD-SLAM \\
\hline
V1\_01\_easy & \textbf{0.035} & 0.066 \\
\hline
V1\_02\_medium & \textbf{0.020} & 0.074 \\
\hline
V1\_03\_difficult &  \textbf{0.048} & 0.089 \\
\hline
V2\_01\_easy & 0.037 & - \\
\hline
V2\_02\_medium &  0.035 & - \\
\hline
V2\_03\_difficult & X & - \\
\hline
MH\_01\_easy &	0.035 & - \\
\hline
MH\_02\_easy & 0.018 & - \\
\hline
MH\_03\_medium & 0.028 & - \\
\hline
MH\_04\_difficult & 0.119 & - \\
\hline
MH\_05\_difficult &  0.060 & - \\
\hline
\end{tabular}
\end{center}
\end{table}

\subsection{Timing Results}
In order to complete the evaluation of the proposed system, we present in Table \ref{tb:times} timing results in three sequences with different image resolutions 
and sensors. The mean and two standard deviation ranges are shown for each thread task. As these sequences contain one single loop, the full BA and some
tasks of the loop closing thread are executed just once and only a single time measurement is reported.The average tracking time per frame is below the inverse of
the camera frame-rate for each sequence, meaning that our system is able to work in real-time.  As ORB extraction in stereo images is parallelized, it can be seen that extracting 
1000 ORB features in the stereo WVGA images of V2\_02 is similar to extracting the same amount
of features in the single VGA image channel of fr3\_office.

The number of keyframes in the loop is shown as reference for the times related
to loop closing. While the loop in KITTI 07 contains more keyframes, the covisibility graph built for the indoor fr3\_office is denser and therefore the loop fusion, pose-graph optimization
and full BA tasks are more expensive. The higher density of the covisibility graph makes the local map contain more keyframes and points and therefore local map tracking and local BA are 
also more expensive.

\begin{table}[t] 
\caption{TUM RGB-D Dataset. Comparison of Translation RMSE (${m}$).}
\label{tb:tum}
\begin{center}
\begin{tabular}{|l|c|c|c|c|c|}
\hline
 \multirow{2}{*}{Sequence}      & ORB-SLAM2  & Elastic- & \multirow{2}{*}{Kintinuous}  & DVO & RGBD \\
         &  (RGB-D) & Fusion &   & SLAM &  SLAM \\
\hline
fr1/desk &  \textbf{0.016} & 0.020 & 0.037 &   0.021 & 0.026 \\
\hline
fr1/desk2 &  \textbf{0.022} & 0.048 & 0.071  & 0.046 & -\\
\hline
fr1/room & 0.047 & 0.068 & 0.075 &   \textbf{0.043} & 0.087 \\
\hline
fr2/desk & \textbf{0.009} & 0.071 & 0.034 &   0.017 & 0.057\\
\hline
fr2/xyz &  \textbf{0.004} & 0.011 & 0.029 &  0.018 & - \\
\hline
fr3/office &\textbf{0.010} & 0.017 & 0.030 &   0.035 & - \\
\hline
fr3/nst &  0.019 & \textbf{0.016} & 0.031 &  0.018 & - \\
\hline
\end{tabular}
\end{center}
\end{table}

\begin{table*}[t]
\caption{Timing Results of Each Thread in Miliseconds (mean $\pm$ 2 std. deviations).}
\label{tb:times}
 \begin{center}
  \begin{tabular}{ | c | l | c | c | c | }
    \hline
    \parbox[t]{3mm}{\multirow{6}{*}{\rotatebox[origin=c]{90}{Settings}}}& Sequence             & V2\_02           & 07                & fr3\_office          \\ \cline{2-5}
    & Dataset              & EuRoC            & KITTI             & TUM                  \\ \cline{2-5}
    & Sensor               & Stereo           & Stereo            & RGB-D                \\ \cline{2-5}
    & Resolution           & $752 \times 480$ & $1226 \times 370$ & $640 \times 480$     \\ \cline{2-5}
    & Camera FPS           & 20Hz             & 10Hz              & 30Hz                 \\ \cline{2-5}
    & ORB Features         & 1000             & 2000              & 1000                 \\ \hline
    \hline
    \parbox[t]{3mm}{\multirow{7}{*}{\rotatebox[origin=c]{90}{Tracking}}}& Stereo Rectification &  $3.43  \pm 1.10$    &     -             &   -   \\ \cline{2-5}
    & ORB Extraction       &  $13.54 \pm 4.60$    &   $24.83 \pm 8.28$                &   $11.48 \pm 1.84$                   \\ \cline{2-5}
    & Stereo Matching      &  $11.26 \pm 6.64$   &   $15.51 \pm 4.12$                &   $0.02 \pm 0.00$                   \\ \cline{2-5}
    & Pose Prediction      &  $2.07 \pm 1.58$    &   $2.36 \pm 1.84$                &   $2.65 \pm 1.28$                   \\ \cline{2-5}
    & Local Map Tracking   &  $10.13 \pm 11.40$   &   $5.38 \pm 3.52$                &   $9.78 \pm 6.42$                   \\ \cline{2-5}
    & New Keyframe Decision&  $1.40 \pm 1.14$    &   $1.91 \pm 1.06$                &   $1.58 \pm 0.92$                   \\ \cline{2-5}
    & Total                &  $41.66 \pm 18.90$   &   $49.47 \pm 12.10$                &   $25.58 \pm 9.76$                   \\ \hline
    \hline
    \parbox[t]{3mm}{\multirow{6}{*}{\rotatebox[origin=c]{90}{Mapping}}}& Keyframe Insertion   &   $10.30 \pm 7.50$    &  $11.61 \pm 3.28$                 &   $11.36 \pm 5.04$                   \\ \cline{2-5}
    & Map Point Culling    &   $0.28 \pm 0.20$     &  $0.45 \pm 0.38$                 &   $0.25 \pm 0.10$                   \\ \cline{2-5}
    & Map Point Creation   &   $40.43 \pm 36.10$   &  $47.69 \pm 29.52$                 &   $53.99 \pm 23.62$                   \\ \cline{2-5}
    & Local BA             &   $137.99 \pm 248.18$ &  $69.29 \pm 61.88$                 &   $196.67 \pm 213.42$                   \\ \cline{2-5}
    & Keyframe Culling     &   $3.80 \pm 8.20$     &  $0.99 \pm 0.92$                 &   $6.69 \pm 8.24$                   \\ \cline{2-5}
    & Total                &   $174.10 \pm 278.80$ &  $129.52 \pm 88.52$                 &   $267.33 \pm 245.10$                   \\ \hline
    \hline
    \parbox[t]{3mm}{\multirow{5}{*}{\rotatebox[origin=c]{90}{Loop}}}& Database Query       &   $3.57 \pm 5.86$     &  $4.13 \pm 3.54$                 &   $2.63 \pm 2.26$                   \\ \cline{2-5}
    & SE3 Estimation       &   $0.69 \pm 1.82$     &  $1.02 \pm 3.68$                 &   $0.66 \pm 1.68$                   \\ \cline{2-5}
    & Loop Fusion          &   $21.84 $            &  $82.70 $                 &   $298.45 $                   \\ \cline{2-5}
    & Essential Graph Opt. &  $73.15$              &  $178.31 $                 &   $281.99 $                   \\ \cline{2-5}
    & Total  &  $108.59 $            &  $284.88 $                 &   $598.70 $                   \\ \hline
    \hline
    \parbox[t]{3mm}{\multirow{3}{*}{\rotatebox[origin=c]{90}{BA}}}& Full BA              &  $349.25 $            &   $1144.06 $                &   $1640.96 $                   \\ \cline{2-5}
    & Map Update           &  $3.13 $              &   $11.82 $                &   $5.62 $                   \\ \cline{2-5}
    & Total                &  $396.02 $            &   $1205.78$                &   $1793.02 $                   \\ \hline
    \hline
    \multicolumn{2}{|c|}{Loop size (\#keyframes)} & 82 & 248 & 225 \\ \hline
  \end{tabular}
\end{center}
\end{table*}

\section{Conclusion} \label{sec:conclusion}

We have presented a full SLAM system for monocular, stereo
and RGB-D sensors, able to perform relocalization, loop closing and
reuse its map in real-time on standard CPUs. We focus on building
globally consistent maps for reliable and long-term localization in a
wide range of environments as demonstrated in the experiments.
The proposed localization mode with the relocalization capability of the system yields a very robust, zero-drift, and ligthweight localization method for known environments. This mode can be
useful for certain applications, such as tracking the user viewpoint
in virtual reality in a well-mapped space.

The comparison to the state-of-the-art shows that ORB-SLAM2 achieves in most cases 
the highest accuracy. In the KITTI visual odometry benchmark ORB-SLAM2 is currently 
the best stereo SLAM solution. Crucially, compared with the stereo visual odometry
methods that have flourished in recent years, ORB-SLAM2 achieves zero-drift localization 
in already mapped areas.

Surprisingly our RGB-D results demonstrate that if the most accurate
camera localization is desired, bundle adjustment performs better
than direct methods or ICP, with the additional advantage of being
less computationally expensive, not requiring GPU processing to
operate in real-time.

We have released the source code of our system,
with examples and instructions so that it can be easily used by
other researchers.
ORB-SLAM2 is to the best of our knowledge the first open-source visual SLAM system 
that can work either with monocular, stereo and RGB-D inputs. Moreover our source code contains 
an example of an augmented reality application\footnote{\url{https://youtu.be/kPwy8yA4CKM}}
using a monocular camera to show the 
potential of our solution.

Future extensions might include, to name some examples, non-overlapping multi-camera, fisheye or omnidirectional
cameras support, large scale dense fusion, cooperative mapping or increased motion blur robustness.
 
\IEEEtriggeratref{12}


\begin{thebibliography}{10}
\providecommand{\url}[1]{#1}
\csname url@rmstyle\endcsname
\providecommand{\newblock}{\relax}
\providecommand{\bibinfo}[2]{#2}
\providecommand\BIBentrySTDinterwordspacing{\spaceskip=0pt\relax}
\providecommand\BIBentryALTinterwordstretchfactor{4}
\providecommand\BIBentryALTinterwordspacing{\spaceskip=\fontdimen2\font plus
\BIBentryALTinterwordstretchfactor\fontdimen3\font minus
  \fontdimen4\font\relax}
\providecommand\BIBforeignlanguage[2]{{%
\expandafter\ifx\csname l@#1\endcsname\relax
\typeout{** WARNING: IEEEtran.bst: No hyphenation pattern has been}%
\typeout{** loaded for the language `#1'. Using the pattern for}%
\typeout{** the default language instead.}%
\else
\language=\csname l@#1\endcsname
\fi
#2}}

\bibitem{MurTRO15}
R.~Mur-Artal, J.~M.~M. Montiel, and J.~D. Tard\'os, ``{ORB-SLAM}: a versatile
  and accurate monocular {SLAM} system,'' \emph{IEEE Trans. Robot.}, vol.~31,
  no.~5, pp. 1147--1163, 2015.

\bibitem{KITTI}
A.~Geiger, P.~Lenz, C.~Stiller, and R.~Urtasun, ``Vision meets robotics: The
  {KITTI} dataset,'' \emph{Int. J. Robot. Res.}, vol.~32, no.~11, pp.
  1231--1237, 2013.

\bibitem{tumrgbd}
J.~Sturm, N.~Engelhard, F.~Endres, W.~Burgard, and D.~Cremers, ``A benchmark
  for the evaluation of {RGB-D SLAM} systems,'' in \emph{IEEE/RSJ Int. Conf.
  Intell. Robots and Syst. (IROS)}, 2012, pp. 573--580.

\bibitem{KinectFusion}
R.~A. Newcombe, A.~J. Davison, S.~Izadi, P.~Kohli, O.~Hilliges, J.~Shotton,
  D.~Molyneaux, S.~Hodges, D.~Kim, and A.~Fitzgibbon, ``{KinectFusion}:
  Real-time dense surface mapping and tracking,'' in \emph{IEEE Int. Symp. on
  Mixed and Augmented Reality (ISMAR)}, 2011.

\bibitem{pazTRO08}
L.~M. Paz, P.~Pini{\'e}s, J.~D. Tard{\'o}s, and J.~Neira, ``Large-scale 6-{DOF}
  {SLAM} with stereo-in-hand,'' \emph{IEEE Trans. Robot.}, vol.~24, no.~5, pp.
  946--957, 2008.

\bibitem{InverseDepth}
J.~Civera, A.~J. Davison, and J.~M.~M. Montiel, ``Inverse depth parametrization
  for monocular {SLAM},'' \emph{IEEE Trans. Robot.}, vol.~24, no.~5, pp.
  932--945, 2008.

\bibitem{WhyFilter}
H.~Strasdat, J.~M.~M. Montiel, and A.~J. Davison, ``Visual {SLAM}: Why
  filter?'' \emph{Image and Vision Computing}, vol.~30, no.~2, pp. 65--77,
  2012.

\bibitem{DWO}
H.~Strasdat, A.~J. Davison, J.~M.~M. Montiel, and K.~Konolige, ``Double window
  optimisation for constant time visual {SLAM},'' in \emph{IEEE Int. Conf.
  Comput. Vision (ICCV)}, 2011, pp. 2352--2359.

\bibitem{RSLAM}
C.~Mei, G.~Sibley, M.~Cummins, P.~Newman, and I.~Reid, ``{RSLAM}: A system for
  large-scale mapping in constant-time using stereo,'' \emph{Int. J. Comput.
  Vision}, vol.~94, no.~2, pp. 198--214, 2011.

\bibitem{sptam}
T.~Pire, T.~Fischer, J.~Civera, P.~De~Crist{\'o}foris, and J.~J. Berlles,
  ``Stereo parallel tracking and mapping for robot localization,'' in
  \emph{IEEE/RSJ Int. Conf. Intell. Robots and Syst. (IROS)}, 2015, pp.
  1373--1378.

\bibitem{StereoLSD}
J.~Engel, J.~Stueckler, and D.~Cremers, ``Large-scale direct {SLAM} with stereo
  cameras,'' in \emph{IEEE/RSJ Int. Conf. Intell. Robots and Syst. (IROS)},
  2015.

\bibitem{Kintinuous}
T.~Whelan, M.~Kaess, H.~Johannsson, M.~Fallon, J.~J. Leonard, and J.~McDonald,
  ``Real-time large-scale dense {RGB-D SLAM} with volumetric fusion,''
  \emph{Int. J. Robot. Res.}, vol.~34, no. 4-5, pp. 598--626, 2015.

\bibitem{rgbdslam}
F.~Endres, J.~Hess, J.~Sturm, D.~Cremers, and W.~Burgard, ``{3-D} mapping with
  an {RGB-D} camera,'' \emph{IEEE Trans. Robot.}, vol.~30, no.~1, pp. 177--187,
  2014.

\bibitem{DVOSLAM}
C.~Kerl, J.~Sturm, and D.~Cremers, ``Dense visual {SLAM} for {RGB-D} cameras,''
  in \emph{IEEE/RSJ Int. Conf. Intell. Robots and Syst. (IROS)}, 2013.

\bibitem{ElasticFusion}
T.~Whelan, R.~F. Salas-Moreno, B.~Glocker, A.~J. Davison, and S.~Leutenegger,
  ``{ElasticFusion}: Real-time dense {SLAM} and light source estimation,''
  \emph{Int. J. Robot. Res.}, vol.~35, no.~14, pp. 1697--1716, 2016.

\bibitem{dorian}
D.~G\'alvez-L\'opez and J.~D. Tard\'os, ``Bags of binary words for fast place
  recognition in image sequences,'' \emph{IEEE Trans. Robot.}, vol.~28, no.~5,
  pp. 1188--1197, 2012.

\bibitem{ORB}
E.~Rublee, V.~Rabaud, K.~Konolige, and G.~Bradski, ``{ORB}: an efficient
  alternative to {SIFT} or {SURF},'' in \emph{IEEE Int. Conf. Comput. Vision
  (ICCV)}, 2011, pp. 2564--2571.

\bibitem{MurICRA14}
R.~Mur-Artal and J.~D. Tard\'os, ``Fast relocalisation and loop closing in
  keyframe-based {SLAM},'' in \emph{IEEE Int. Conf. on Robot. and Autom.
  (ICRA)}, 2014, pp. 846--853.

\bibitem{g2o}
R.~Kuemmerle, G.~Grisetti, H.~Strasdat, K.~Konolige, and W.~Burgard, ``g2o: A
  general framework for graph optimization,'' in \emph{IEEE Int. Conf. on
  Robot. and Autom. (ICRA)}, 2011, pp. 3607--3613.

\bibitem{HaukeScale}
H.~Strasdat, J.~M.~M. Montiel, and A.~J. Davison, ``Scale drift-aware large
  scale monocular {SLAM}.'' in \emph{Robot.: Sci. and Syst. (RSS)}, 2010.

\bibitem{euroc}
M.~Burri, J.~Nikolic, P.~Gohl, T.~Schneider, J.~Rehder, S.~Omari, M.~W.
  Achtelik, and R.~Siegwart, ``The {EuRoC} micro aerial vehicle datasets,''
  \emph{Int. J. Robot. Res.}, vol.~35, no.~10, pp. 1157--1163, 2016.

\bibitem{ORBIMU}
R.~Mur-Artal and J.~D. Tard\'os, ``Visual-inertial monocular {SLAM} with map
  reuse,'' \emph{IEEE Robot. and Autom. Lett.}, vol.~2, no.~2, pp. 796 -- 803,
  2017.


\end{thebibliography}
\end{document}